\pdfoutput=1

\documentclass[11pt]{article}

\usepackage[final]{acl}
\usepackage{times}
\usepackage{latexsym}
\usepackage{hyperref}
\usepackage{amsmath}
\usepackage{amsfonts}
\usepackage[T1]{fontenc}

\usepackage[utf8]{inputenc}
\usepackage{booktabs}
\usepackage{microtype}
\usepackage{graphicx}
\usepackage{dashrule}
\usepackage{amsfonts,amssymb}
\usepackage{arydshln} 
\usepackage{booktabs}
\usepackage{multirow}
\usepackage{subcaption}
\usepackage{microtype}
\newcommand{\eg}{\textit{e}.\textit{g}.\ }
\title{Structural Characterization for Dialogue Disentanglement}

\author{
Xinbei Ma\textsuperscript{\rm 1,2},
Zhuosheng Zhang\textsuperscript{\rm 1,2},
Hai Zhao\textsuperscript{\rm 1,2,\thanks{\; Corresponding author. This paper was partially supported by Key Projects of National Natural Science Foundation of China (U1836222 and 61733011).}} \\
  $^1$Department of Computer Science and Engineering,  Shanghai Jiao Tong University
  \\ $^2$Key Laboratory of Shanghai Education Commission for Intelligent Interaction \\
and Cognitive Engineering, Shanghai Jiao Tong University \\
  \texttt{sjtumaxb@sjtu.edu.cn,zhangzs@sjtu.edu.cn,zhaohai@cs.sjtu.edu.cn} 
  }
\begin{document}
\maketitle
\begin{abstract}
Tangled multi-party dialogue contexts lead to challenges for dialogue reading comprehension, where multiple dialogue threads flow simultaneously within a common dialogue record, increasing difficulties in understanding the dialogue history for both human and machine. 
Previous studies mainly focus on utterance encoding methods with carefully designed features but pay inadequate attention to characteristic features of the structure of dialogues. 
We specially take structure factors into account and design a novel model for dialogue disentangling. 
Based on the fact that dialogues are constructed on successive participation and interactions between speakers, we model structural information of dialogues in two aspects: 1)speaker property that indicates whom a message is from, and 2) reference dependency that shows whom a message may refer to. The proposed method achieves new state-of-the-art on the Ubuntu IRC benchmark dataset and contributes to dialogue-related comprehension.
\end{abstract}

\section{Introduction}
\begin{figure}[t]
		\centering
		\includegraphics[width=0.48\textwidth]{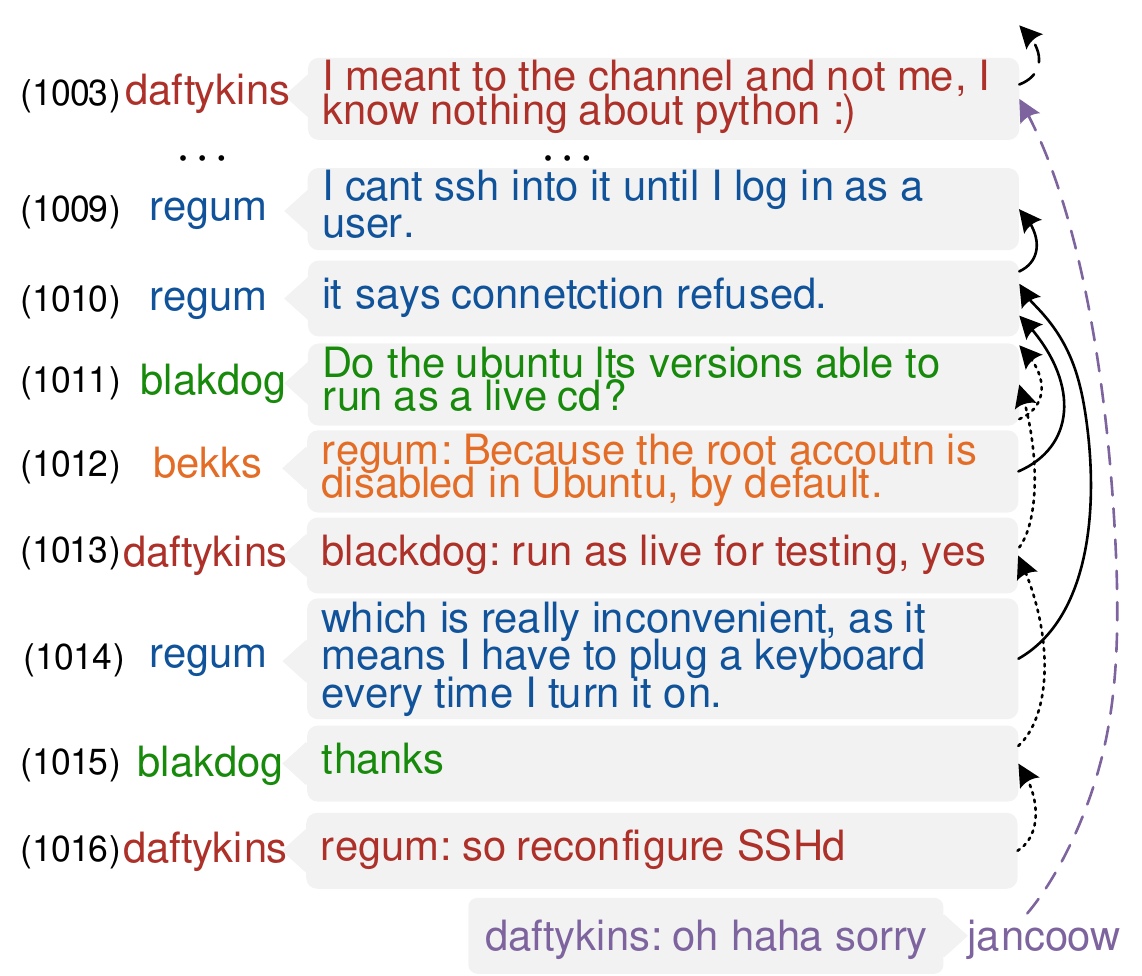}
		\caption{\label{example} Here is an example piece of multi-party chatting logs from Ubuntu IRC \cite{kummerfeld2019large}. \texttt{jancoow} figures out conversation threads, understand the context and reply to the related message (\texttt{1003} from \texttt{daftykins}), and the dialogue develops.}
\end{figure}

Communication between multiple parties happens anytime and anywhere, especially as the booming social network services hugely facilitate open conversations, such as group chatting and forum discussion, producing various tangled dialogue logs \cite{Loweubuntu,Zhangedc,choi2018quac,reddy2019coqa, li2020molweni}.
Whereas, it can be challenging for a new participant to understand the previous chatting log since multi-party dialogues always exhibit disorder and complication \cite{shen2006thread, elsner2010disentangling, jiang-etal-2018-learning-disentangle, kummerfeld2019large}. In fact, it is because of the distributed and random organization, multi-party dialogues are much less coherent or consistent than plain texts.
As the example shown in figure \ref{example}, the development of a multi-party dialogue has the following characteristics: 1) Random users successively participate in the dialogue and follow specific topics that they are interested in, motivating the development of those topics. 2) Users reply to former related utterances and mention involved users, forming dependencies among utterances. 
As a result, multiple ongoing conversation threads grow as the dialogue proceeds, which breaks the consistency and hinders both humans and machines from understanding the context, let alone giving a proper response \cite{jiang-etal-2018-learning-disentangle,kummerfeld2019large,joty2019discourse, jiang2021dialogue}.
In a word, the behavior of speakers determines the structure of a dialogue passage. And the structure causes problems of reading comprehension. Hence, for better understanding, structural features of dialogue context deserve special attention.

Disentanglement is worthy of study.
Decoupling messages or clustering conversation threads help with screening concerned parts among contexts, therefore it may be naturally required by passage comprehension, and related downstream dialogue tasks \cite{elsner2010disentangling, jia2020multi,liu2021unsupervised}, such as response selection, question-answering, etc. 

Nevertheless, existing works on dialogue disentanglement \cite{zhu2020did, yu2020online, li2020dialbert} generally ignore or pay little attention to characters of dialogues. 
Earlier works mainly depend on feature engineering \cite{kummerfeld2019large, elsner2010disentangling, yu2020online}, and use well-constructed handcrafted features to train a naive classifier \cite{elsner2010disentangling} or linear feed-forward network \cite{kummerfeld2019large}. 
Recent works are mostly based on two strategies: 1) two-step \cite{mehri-carenini-2017-chat, zhu2020did, yu2020online, li2020dialbert,liu2021unsupervised} and 2) end-to-end \cite{tan-etal-2019-context,liu2020end}. 
In terms of the two-step method, the disentanglement task is divided into \textit{matching} and \textit{clustering}. It means firstly matching utterance pairs to detect reply-to relations and then clustering utterances according to the matching score. 
In the end-to-end strategy, alternatively, for each conversation thread, the state of dialogue is modeled, and is mapped with a subsequent utterance to update. At the same time, the subsequent utterance is judged to belong to the best-matched thread. 
Nonetheless, the essence of both strategies is to model the relations of utterance pairs.

Recently, Pre-trained Language Models (PrLMs) \cite{devlin2019bert, YinhanLiuroberta, ClarkLLM20} have brought prosperity to numbers of natural language processing tasks by providing contextualized backbones. Various works have reported substantial performance gains with the contextualized information from PrLMs \cite{Loweubuntu, li2020molweni, liumdfn,jia2020multi,WangHJ20}. 
Studies on dialogue disentanglement also get benefit from PrLMs \cite{li2020dialbert, zhu2020did}, whereas, there is still room for improvement due to their insufficient enhancement of dialogue structure information.

So as to enhance characteristic structural features of tangled multi-party dialogues, we design a new model as a better solution for dialogue disentanglement.
Structure of a multi-party dialogue is based on the actions of speakers according to the natural development of dialogues. 
Hence, we model two structural features to help with the detection of reply-to relationships: 1)user identities of messages, referred to as \textit{speaker property}; and 2) mention of users in messages, called \textit{reference dependency}. With the two features enhanced between encoding and prediction, the model makes progress on dialogue disentanglement.
Evaluation is conducted on DSTC-8
Ubuntu IRC dataset \cite{kummerfeld2019large}, where our proposed model achieves new state-of-the-art.
Further analyses and applications illustrate the advantages and scalability additionally. 
Our source code is available \footnote{\url{https://github.com/xbmxb/StructureCharacterization4DD}}.

\section{Background and Related Work}


\subsection{Dialogue-related Reading Comprehension}
Dialogue understanding brings challenges to machine reading comprehension (MRC), in terms of handling the complicated scenarios from multiple speakers and criss-crossed dependencies among utterances \cite{Loweubuntu,yang-choi-2019-friendsqa,sun2019dream, li2020molweni}.
A dialogue is developed by all involved speakers in a distributed way. An individual speaker focuses on some topics that are discussed in the conversation, and then declares oneself or replies to utterances from related speakers. 
Therefore, consistency and continuity are broken by tangled reply-to dependencies between non-adjacent utterances \cite{li2020molweni, jia2020multi, ma2021enhanced, li2021dadgraph}, leading to a graph structure that is different from smooth presentation in plain texts.

PrLMs have made a significant breakthrough in MRC,
where various training objectives and strategies \cite{devlin2019bert,ClarkLLM20, YinhanLiuroberta, albert/iclr/LanCGGSS20} have achieved further improvement.
Devoted to MRC tasks, PrLMs usually work as a contextualized encoder with some task-oriented decoders added \cite{devlin2019bert}. And this paradigm may be a generic but suboptimal solution, especially for some distinctive scenarios, such as dialogue.

Recently, numbers of works of dialogue-related MRC have managed to enhance dialogue structural features in order to deal with dialogue passages better \cite{liumdfn, jia2020multi, zhang-zhao-2021-structural, ma2021enhanced, li2021dadgraph}, which achieve progress compared to methods that were previously proposed for plain texts. 
This inspiration impacts and promotes a wide range of dialogue-related MRC tasks such as response selection \cite{Gusabert,liumdfn}, question answering \cite{ma2021enhanced, li2021dadgraph}, emotion detection \cite{HuWH20}, etc.

\subsection{Dialogue Disentanglement}
Dialogue disentanglement \cite{elsner2010disentangling}, which is also referred to as conversation management \cite{traum2004evaluation} , thread detection \cite{shen2006thread} or thread extraction \cite{adam-calhoun}, has been studied for decades, 
since understanding long multi-party dialogues remains to be non-trivial.
Thus, dialogue disentanglement methods have been proposed to cluster utterances.

Early works can be summarized as feature encoder and clustering algorithms.
Well-designed handcraft features are constructed as input of simple networks that predict whether a pair of utterances are alike or different, and clustering methods are then borrowed for partitioning \cite{elsner2010disentangling, jiang-etal-2018-learning-disentangle}. 
Researches are facilitated by a large-scale, high-quality public dataset, Ubuntu IRC, created by \citet{kummerfeld2019large}. And then the application of FeedForward network and pointer network \cite{vinyals2015pointer} leads to significant progress, but the improvement still partially relies on handcraft-related features \cite{kummerfeld2019large, yu2020online}.
Then the end-to-end strategy is proposed and fills the gap between the \textit{match} and \textit{clustering} \cite{liu2020end}, where dialogue disentanglement is modeled as a dialogue state transition process. The utterances are clustered by mapping with the states of each dialogue thread.
Inspired by achievements of pre-trained language models \cite{devlin2019bert,ClarkLLM20, YinhanLiuroberta}, latest work use BERT to contextually encode the dialogue context \cite{zhu2020did, li2020dialbert}.
\citet{DBLP:liu/emnlp/0033SZ21} investigates disentanglement from a different perspective. Their end-to-end co-training approach provides a novel unsupervised baseline.

However, attention paid to the characteristics of dialogues seems to be inadequate. 
Feature engineering-based works represent properties of individual utterances such as time, speakers, and topics with naive handcraft methods, thus ignoring dialogue contexts \cite{elsner2010disentangling, kummerfeld2019large}. 
PrLM-based Masked Hierarchical Transformer \cite{zhu2020did} utilizes the golden conversation structures to operate attentions on related utterances when training models, which results in exposure bias. 
DialBERT \cite{li2020dialbert}, a recent architecture including a BERT \cite{devlin2019bert} and an LSTM \cite{hochreiter1997lstm}, models contextual clues but no dialogue-specific features, and claims a state-of-the-art performance. 
Our approach draws inspiration from these works and further models structural features for better dialogue understanding.

Unlike the above studies, our work incorporates dialogue-specific characters.
We propose a new model considering structural characteristics of dialogues, based on the fact that dialogues are developed according to the behavior of speakers.
In detail, we model dialogue structures with two highlights: 
1) speaker properties of each utterance and 2) reference of speakers between utterances, which both help with modeling inherent interactions among a dialogue passage.


\subsection{Speaker-aware Dialogue Modeling}
Speaker role, as a feature of dialogue passage, has received growing attention recently. 
On the one hand, speaker embedding facilities research of dialogues.
Speaker-aware modeling has also made contributions to response retrieval \cite{Gusabert, liumdfn}. SA-BERT \cite{Gusabert} add a speaker embedding to the input of a PrLM, while MDFN \cite{liumdfn} modifies self-attention to enhance speaker switches.
Persona has been utilized for smoother dialogue generation. In recent work \cite{liu2020personachat}, the speaker-aware information is modeled by adding a reward of persona proximity to the reinforcement learning of generation, based on a persona-annotated dataset \cite{zhang-etal-2018-personalizing}. 
On the other hand, speakers role is a valuable research object for personal knowledge analysis, since the persona can be extracted from one's words in dialogues. 
Relationship prediction task has been better handled through observing interactions of dialogue speakers \cite{JiaHZ21, tigunova-etal-2021-pride}. \citet{tigunova-etal-2021-pride} make use of speaker identity by a \textit{SA-BERT} \cite{Gusabert}-like embedding but in utterance-level representation.

Relations between utterances have been studied for a long time. Earlier works mostly based on pioneer datasets, Penn Discourse TreeBank \cite{prasad-etal-2008-penn} and Rhetorical Structure Theory Discourse TreeBank \cite{mann1988rhetorical}. In the dialogue field, the much more complex relations contain latent features \cite{shi2019deep, zhang-zhao-2021-structural, jia2020multi}. Due to the inherent graph structure, Graph Convolutional Network \cite{KipfW17} is well applied to natural language modeling. Derivations such as Relational-GCN \cite{schlichtkrull2017modeling}, TextGCN \cite{DBLP:conf/aaai/YaoM019}, LBGCN \cite{huang-etal-2021-lexicon-based}, etc, encourage better structural solutions in NLP.

In this work, we aim to inject speaker-aware and reference-aware characteristic features for the motivation of disentanglement, instead of making progress on embedding approaches.

\section{Methodology}
The definition of the dialogue disentanglement task and details of our model are sequentially presented in this section, illustrating how we make efforts for disentanglement with dialogue structural features.

\begin{figure*}[htb]
	\centering
	\includegraphics[width=\textwidth]{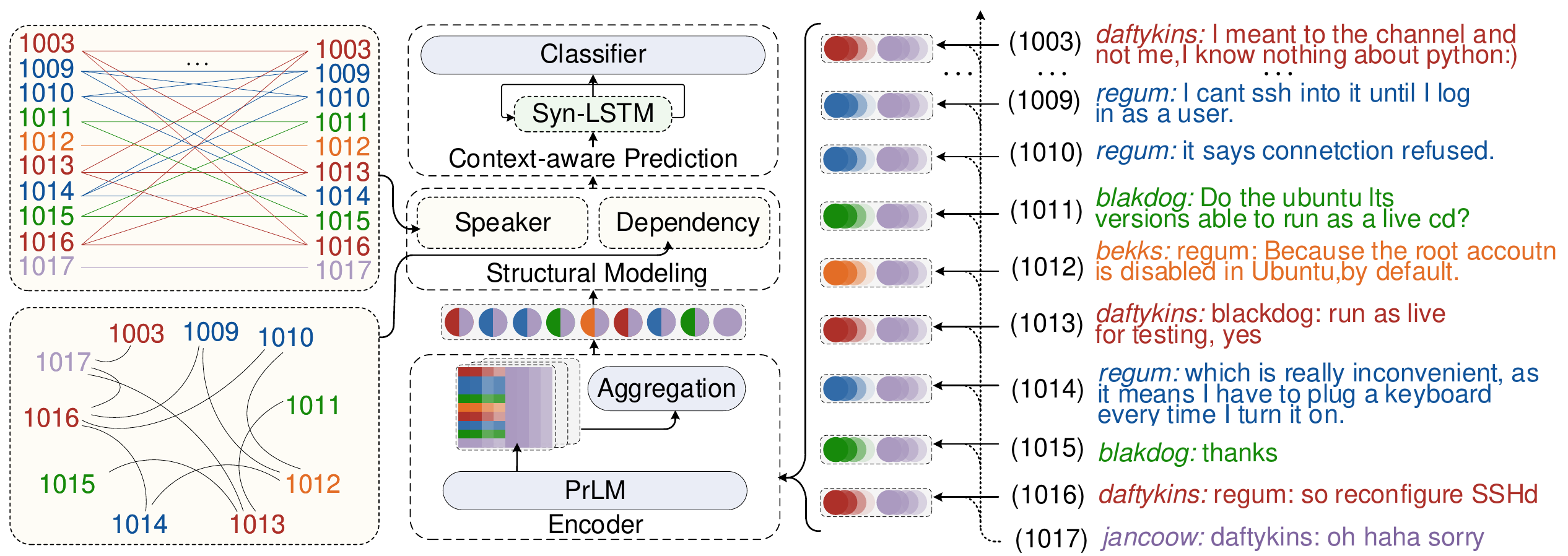}
	\caption{Overview of the model and data flow. A dialogue is encoded to context-level in the encoder module. Then speaker-aware and reference-aware features are enhanced in the structural modeling layer. And context-aware prediction model makes the final prediction.}
	\label{overview}
\end{figure*}

\subsection{Task Formulation}
Suppose that we perform disentanglement to a long multi-party dialogue history $\mathbb{D}=\{u_0,u_2,\dots,u_n\}$, where $\mathbb{D}$ is composed of $n$ utterances. An utterance includes an identity of speaker and a message sent by this user, thus denoted as $u_i=\{s_i, m_i\}$. As several threads are flowing simultaneously within $\mathbb{D}$, we define a set of threads $\mathbb{T}=\{t_0,t_2,\dots,t_p\}$ as a partition of $\mathbb{D}$, where $t_i=\{u_{i_0},\dots,u_{i_k}\}$ denoting a thread of the conversation. In this task, we aim to disentangle  $\mathbb{D}$ into $\mathbb{T}$.
As indicated before, a multi-party dialogue is constructed by successive participation of speakers, who often reply to former utterances of interest. Thus, a dialogue passage can be modeled as a graph structure whose vertices denote utterances and edges denote reply-to relationships between utterances. Following the two-step method \cite{mehri-carenini-2017-chat}, we focus on finding a parent node for each utterance through inference of reply-to relationship, so as to discover edges and then determine the graph of a conversation thread.

\subsection{Model Architecture}
Figure \ref{overview} shows the architecture of the proposed model, which is introduced in detail in this part. The model architecture consists of three modules, including text encoder, structural interaction, and context-aware prediction:
1) The utterances from a dialogue history are encoded with a PrLM, whose output is then aggregated to context-level. 
2) The representation is sequentially fed into the structural modeling module, where dialogue structural features are used to characterize contexts.
3) Then in the prediction module, the model performs a fusion and calculates the prediction of reply-to relationships.

\subsubsection{Encoder}
\textbf{Pairwise encoding}
\label{sec:sacrify}
Following previous works \cite{zhu2020did,li2020dialbert}, we utilize a pre-trained language model \eg BERT \cite{devlin2019bert} as an encoder for contextualized representation of tokens. Since chatting records are always long and continuous, it is inappropriate and unrealistic to concatenate the whole context as input. Hence, we focus on the pair of utterances with a reply-to relation. An utterance is concatenated with each parent candidate as input to a PrLM. This may sacrifice contextual information between candidates, but we make up for this in \ref{sec:makeup}. 

Assuming that for an utterance $u_i$, we consider former $C$ utterances (including $u_i$ itself) as candidates for parent node of $u_i$, the input of a PrLM is in the form of \texttt{[CLS] $u_{i-j}$ [SEP] $U_i$ [SEP]}, where $0 \leq j \leq C-1$. The output is denoted as $H_0 \in \mathbb{R}^{C\times L\times D}$, where $C$ denotes the window length in which former utterances are considered as candidates of the parent, $L$ denotes the input sequence length in tokens, $D$ denotes the dimension of hidden states of the PrLM.
Note that there is a situation where the golden parent utterance is beyond the range of $[u_{i-(C-1)}, u_i]$. We label a self-loop for $u_i$ in this case, which means being too far from the parent making $u_i$ a beginning of a new dialogue thread. 
It makes sense in the real world, because when users join in a chat (\eg entering a chatting room), they intend to check a limited number of recent messages and make replies, instead of scanning the entire chatting record.

\noindent
\textbf{Utterance Aggregation}
$H_0$ is pairwise contextualized representations of each pair of token sequences $(u_{i-j}, u_i)$, thus need to be aggregated to context-level representation for further modeling. 
Since special token \texttt{[CLS]} makes more sense on classification tasks \cite{devlin2019bert},
we simply reserve the representations of \texttt{[CLS]}.
The concatenated pairwise context-level representations from all candidates is denoted as $H_1 \in \mathbb{R}^{C\times D}$, where $C$ denotes the window length and $D$ denotes the dimension of hidden states of the PrLM.

\subsubsection{Structural Modeling}
For our structural modeling, a simple but effective method is preferred. Hence, for speaker property, we applied the idea of masked MHSA method \cite{liumdfn} for better effectiveness and conciseness \cite{ma2021enhanced}.
In dependency modeling, we only built one relation type, i.e., reference, where a vanilla r-GCN \cite{schlichtkrull2017modeling} is an appropriate baseline method. 

\noindent
\textbf{Speaker Property Modeling}
We use the term \textit{Speaker Property} to denote the user identity from whom an utterance is, in formulation, $s_i$. Modeling speaker property could be worthwhile because sometimes a participant may focus on conversations with specific speakers.
Following the idea of masking attention \cite{liumdfn}, we build a Multi-Head Self-Attention (MHSA) mechanism to emphasize correlations between utterances from the same speaker.
The mask-based MHSA is formulated as follows:
\begin{equation}
\setlength{\abovedisplayskip}{3pt}
\setlength{\belowdisplayskip}{3pt}
\begin{split}
&\textit{A} \textup{(} \textit{Q, K, V, M} \textup{)} = \textup{softmax} \textup{(} \frac{\textit{QK}^{T}}{\sqrt\textit{d}_{k}} \textup{ $+$ } \textit{M} \textup{)} \textit{V} \textup{, }\\
&\textit{head}_{t} = \textit{A} \textup{(} \textit{HW}_{t}^{Q} \textit{, } \textit{HW}_{t}^{K} \textit{, } \textit{HW}_{t}^{V} \textit{, } \textit{M} \textup{), }\\
&\textit{MHSA} \textup{(} \textit{H, M} \textup{)} = \textup{[} \textit{head}_{1} \textit{, } \textit{, \dots ,} \textit{head}_{N} \textup{]} \textit{W}^{O} \textup{, }\\
\end{split}
\label{eq3—1}
\end{equation}
where $A$, $head_t$, $Q$, $K$, $V$, $M$, $N$ denote the attention, head, query, key, value, mask, and the number of heads, respectively. $H$ denotes the input matrix, and $W_t^Q$, $W_t^K$, $W_t^V$, $W^O$ are parameters. Operator [$\cdot,\cdot$] denotes concatenation. At this stage, the input of MHSA is the aggregated representation $H_1$ with a speaker-aware mask matrix $M$. The element at the $i$-th row, $j$-th column of $M$ depend on speaker properties of $u_i$ and $u_j$:
\begin{equation}
\setlength{\abovedisplayskip}{3pt}
\setlength{\belowdisplayskip}{3pt}
\begin{split}
\textit{M} \textit{[i, j]} &=
\begin{cases}
0,& \textit{s}_{i} \textup{$=$} \textit{s}_{j}\\
-\infty,& \text{otherwise}\\
\end{cases}\\
\textit{H}_{2} &= \textit{MHSA} \textup{(} \textit{H}_{1} \textup{, } \textit{M} \textup{),}\\
\end{split}
\label{eq2}
\end{equation}
The output of MHSA, $H_{MHSA}$, has the same dimension with $H_1 \in \mathbb{R}^{C\times D}$.
We concatenate $H_1$ and $H_{MHSA}$ and adjust to the same size using a linear layer, resulting in an output of this module denoted as $H_2 \in \mathbb{R}^{C\times D}$.

\begin{table*}[t]
	\centering
	\setlength{\tabcolsep}{8pt}\small
	{
	\begin{tabular}{lllllll}  
	\toprule
{\textbf{Model}} 
		 &\textbf{VI} &\textbf{ARI} &\textbf{1-1} &\textbf{F1} &\textbf{P} &\textbf{R} \\
        \midrule
    \multicolumn{7}{c}{\textit{Test Set}}\\
    FeedForward \cite{kummerfeld2019large} & 91.3 &-- &75.6 & 36.2 &34.6 & 38.0 \\
    \quad $\times 10$ union \cite{kummerfeld2019large} & 86.2  & -- &62.5 & 33.4 & 40.4 &28.5 \\
    \quad $\times 10$ vote \cite{kummerfeld2019large} & 91.5  & -- &76.0 & 38.0 & 36.3 &39.7 \\
    \quad $\times 10$ intersect \cite{kummerfeld2019large} & 69.3  & -- &26.6 & 32.1 & 67.0 &21.1 \\
    Elsner \cite{elsner-charniak-2008-talking} & 82.1 & -- &51.4 &15.5 & 12.1 &21.5  \\
    Lowe \cite{lowe2017training} & 80.6 & -- & 53.7 & 8.9 & 10.8 & 7.6\\
    BERT \cite{li2020dialbert} &90.8 &62.9 & 75.0 &32.5 & 29.3 & 36.6 \\
    DialBERT \cite{li2020dialbert} &92.6 &69.6 & 78.5 &44.1 & 42.3 & 46.2\\
    \quad +cov \cite{li2020dialbert} &93.2 & 72.8 & 79.7 & 44.8 &42.1 & 47.9\\
    \quad +feature \cite{li2020dialbert} &92.4 & 66.6 & 77.6 & 42.2 &38.8 & 46.3\\
    \quad +future context \cite{li2020dialbert} &92.3 & 66.3 & 79.1 & 42.6 &40.0 & 45.6\\
    Ptr-Net \cite{yu2020online} & 92.3 &70.2 &--  & 36.0 & 33.0 & 38.9\\
    \quad + Joint train \cite{yu2020online} & 93.1  & 71.3 &--  & 39.7 & 37.2 & 42.5  \\
    \quad + Self-link \cite{yu2020online} & 93.0 &74.3 & --  &41.5 & 42.2 & 44.9 \\
    \quad + Joint train\&Self-link \cite{yu2020online} & 94.2 & \textbf{80.1} & - & 44.5 & 44.9 & 44.2 \\
    \cdashline{1-7}[0.8pt/2pt]
    BERT$_{base}$ (Our baseline) &91.4 & 60.8 & 74.4 & 37.2 &34.0 & 41.2 \\
    Our model &\textbf{94.6}\scriptsize{+3.2} & 76.8\scriptsize{+16} & \textbf{84.2}\scriptsize{+9.8} &  \textbf{51.7}\scriptsize{+14.5} &\textbf{51.8}\scriptsize{+17.8} & \textbf{51.7}\scriptsize{+10.5} \\
    \midrule
    \multicolumn{7}{c}{\textit{Dev Set}}\\
    Decom. Atten. \cite{parikh2016decomposable}&70.3& -- & 39.8 & 0.6 & 0.9 &0.7 \\
    \quad +feature\cite{parikh2016decomposable} & 87.4 & -- & 66.6 & 21.1 & 18.2 & 25.2 \\
    ESIM \cite{chen2017enhanced}&72.1 & -- & 44.0 & 1.4 & 2.2 & 1.8  \\
    \quad +feature \cite{chen2017enhanced}&87.7 & -- & 65.8 & 22.6 & 18.9 & 28.3 \\
    MHT \cite{zhu2020did} & 82.1 & -- & 59.6 & 8.7 & 12.6 &10.3 \\
    \quad +feature \cite{zhu2020did} & 89.8 & -- & 75.4  & 35.8 & 32.7 &  34.2 \\
    DialBERT \cite{li2020dialbert} &94.1 &81.1 & 85.6 &48.0 & 49.5 & 46.6 \\
    \cdashline{1-7}[0.8pt/2pt]
    BERT$_{base}$ (Our baseline) & 92.8 &74.4 & 80.8 & 40.8 &37.7 & 42.7  \\
    Our model & \textbf{94.4}\scriptsize{+1.6} & \textbf{81.8}\scriptsize{+7.4} & \textbf{86.1}\scriptsize{+5.3} & \textbf{52.6}\scriptsize{+11.8} &\textbf{51.0}\scriptsize{+13.3}& \textbf{54.3}\scriptsize{+11.6} \\
	\bottomrule
	\end{tabular}
	}
	\caption{Experimental results on the Ubuntu IRC dataset \cite{kummerfeld2019large}.}
	\label{maintable}
\end{table*}

\noindent
\textbf{Reference Dependency Modeling}
As discussed above, the relation of references between speakers is the most important and straightforward dependency among utterances. Because references indicate interactions between users, it is the internal motivation of the development of a dialogue. To this end, we build a matrix to label the references, which is regarded as an adjacency matrix of a graph representation. In the graph of references, a vertice denotes an utterance and an edge for reference dependence.
For example, $u_{1012}$ in Figure \ref{example} mentions and reply to \textit{regum}, forming dependence to utterances from \textit{regum}, i.e., $u_{1009}$, $u_{1010}$, and $u_{1014}$. Thus there are edges from $v_{1012}$ to $v_{1009}$, $v_{1010}$, and $v_{1014}$. Impressed by the significant influence of graph convolutional network (GCN) \cite{KipfW17}, we borrow the relation-modeling of relational graph convolutional network (r-GCN) \cite{schlichtkrull2017modeling, shi2019deep} in order to enhance the reference dependencies, which can be denoted as follows:
\begin{equation}\nonumber
\setlength{\abovedisplayskip}{3pt}
\setlength{\belowdisplayskip}{3pt}
\textit{h}_{i}^{(l+1)} = \text{$\sigma$($\sum_{r \in \mathcal{B}}\sum_{j\in N_i^r}\frac{1}{c_{i,r}}$} \textit{W}_{r}^{(l)} \textit{h}_{j}^{(l)} \textup{$+$} \textit{W}_{0}^{(l)} \textit{h}_{i}^{(l)} \textup{),}
\label{eq4}
\end{equation}
where $\mathcal{B}$ is the set of relationships, which in our module is only reference dependencies. $N_i^r$ denotes the set of neighbours of vertice $v_i$, which are connected to $v_i$ through relationship $r$, and $c_{i,r}$ is constant for normalization. $W_r^{(l)}$ and $W_0^{(l)}$ are parameter matrix of layer $l$. $\sigma$ is activated function, which in our implementation is ReLU \cite{glorot2011deep, agarap2018deep}.
$H_2$ is fed into this module and derives $H_3 \in \mathbb{R}^{C\times D}$ through r-GCN. 
\subsubsection{Context-aware Prediction}
\label{sec:makeup}
The structure-aware representation $H_3$ needs to be combined with the original representation of \texttt{[CLS]} $H_0$ for enhancement. An LSTM-like layer \cite{hochreiter1997lstm, li2020dialbert} can be utilized for compensating contextualized information of the whole candidate window.

Motivated by the two points above, we employ a Syn-LSTM module \cite{xu2021better}, which was originally proposed for named entity recognition (NER). 
A Syn-LSTM is distinguished from an additional input gate for an extra input source, whose parameters are trainable, achieving a better fusion of two input sources. 
Thus, a layer of Syn-LSTM models the contextual information while the reference dependency is highlighted, enriching relations among parent candidates. 
In a Syn-LSTM cell, the cell state is derived from the two input and former state as well:
\begin{equation}\nonumber
\setlength{\abovedisplayskip}{3pt}
\setlength{\belowdisplayskip}{3pt}
\begin{split}
&\textit{c}_{{1}_{t}} = \textup{tanh(W}^{(k)} \textup{x}_{{1}_{t}} + \textup{U}^{(k)} \textup{h}_{t-1} +\textup{b}_{k} \textup{),}\\
&\textit{c}_{{2}_{t}} = \textup{tanh(W}^{(p)} \textup{x}_{{2}_{t}} + \textup{U}^{(p)} \textup{h}_{t-1} +\textup{b}_{p} \textup{),}\\
&\textit{c}_{t} = \textit{f}_{t} \textup{$\odot$}\textit{c}_{t-1} + \textit{i}_{{1}_{t}} \textup{$\odot$}\textit{c}_{{1}_{t}} + \textit{i}_{{2}_{t}} \textup{$\odot$}\textit{c}_{{2}_{t},}\\
&\textit{h}_{t} = \textit{o}_{t} \textup{$\odot$} \textup{tanh(c}_{t}\textup{),}\\
\end{split}
\label{syn}
\end{equation}
where $f_t, o_t, i_{1_t}, i_{2_t}$ are forget gate, output gate and two input gates. $c_{t-1}, c_t$ denote former and current cell states. $h_{t-1}$ is former hidden state. And $W, U, b$ are learnable parameters. 
We use the Syn-LSTM in a bi-directional way, and the output is denoted as $H_4 \in \mathbb{R}^{C\times2D_{r}}$, where $D_r$ is the hidden size of the Syn-LSTM.

At this stage, $H_4$ is the structural feature-enchanced representation of each pair of the utterance $U_i$ and a candidate parent utterance $u_{i-j}$. To measure the correlations of these pairs, we follow previous work \cite{li2020dialbert} to consider the Siamese architecture between each $[u_i,u_{i-j}]$ pair ($1 \leq j \leq C-1$) and $[u_i,u_i]$ pair:
\begin{equation}\nonumber
\setlength{\abovedisplayskip}{3pt}
\setlength{\belowdisplayskip}{3pt}
\textup{H}_{5}[j] = \textup{[p}_{ii}, \textup{p}_{ij}, \textup{p}_{ii} \odot p_{ij}, \textup{p}_{ii} - \textup{p}_{ij}\textup{]}, 
\label{siamese}
\end{equation}
where $p_{ij}$ is the representation for the pair of $[U_i,U_{i-j}]$ from $H_4$, and we got $H_4 \in \mathbb{R}^{C\times8D_{r}}$. 
$H_5$ is then fed into a classifier to predict the most correlated pair and predict the parent. Cross-entropy loss is used as the model training objective.

\section{Experiments}
Our proposed model is evaluated on a large-scale multi-party dialogue log dataset Ubuntu IRC \cite{kummerfeld2019large}, which is also used as a dataset of DSTC-8 Track2 Task4. The results show that our model surpasses the baseline significantly and achieves a new state-of-the-art.
\subsection{Dataset}
Ubuntu IRC (Internet Relay Chat) \cite{kummerfeld2019large} is the first available dataset and also the largest and most influential benchmark corpus for dialogue disentanglement, which promotes related research heavily.
It is collected from \texttt{\#Ubuntu} and \texttt{\#Linux} IRC channels in the form of chatting logs. The usernames of dialogue participants are reserved, and reply-to relations are manually annotated in the form of \texttt{(parent utterance, son utterance)}.
Table \ref{kumm} shows statics of Ubuntu IRC.

\begin{table}[hbt]
	\centering
	\setlength{\tabcolsep}{6pt}\small
	{\begin{tabular}{lcccc}
		\toprule
		  &Passages & Utterances &Links  &Avg. users \\ 
		\midrule
        \text{Train}  & 153 &22,0463 & 69,395 &130.3 \\
        \text{Dev}  &  10 &12,500 & 2,607 &128.1  \\
        \text{Test}  &  10 &15,000 & 5,187 &156.9 \\
		\bottomrule
	\end{tabular}
	}
	\caption{Statistics of Ubuntu IRC \cite{kummerfeld2019large}.}
	\label{kumm}
\end{table}

\subsection{Metrics}
\paragraph{Reply-to relations} 
We calculate the accuracy for the prediction of parent utterance, indicating the inference ability for reply-to relations.

\paragraph{Disentanglement}
For the goal of dialogue disentanglement, threads of a conversation are formed by clustering all related utterances bridged by reply-to relations, in other words, a connected subgraph. At this stage, we use metrics to evaluate following DSTC-8, which are scaled-Variation of Information (VI) \cite{kummerfeld2019large}, Adjusted rand index (ARI) \cite{hubert1985comparing}, One-to-One Overlap (1-1) \cite{elsner2010disentangling}, precision (P), recall (R), and F1 score of clustering. Note that in the table of results, we present 1-VI instead of VI \cite{kummerfeld2019large}, thus for all metrics, we expect larger numerical values that mean stronger performance.
\subsection{Setup}
Our implementations are based on \textit{Pytorch} and \textit{Transformers} Library \cite{wolf2020transformers}. 
We fine-tune the model employing AdamW \cite{adamw} as the optimizer. The learning rate begins with 4e-6. In addition, due to the trade-off for computing resources, the input sequence length is set to 128, which our inputs are truncated or padded to, and the window width of considered candidates is set to 50.

\subsection{Experimental Results}
As is presented in Table \ref{maintable}, the experimental results show that our model outperforms all baselines by a large margin as the annotated difference values. 
It is also shown that our model achieves superior performance on most metrics compared to previously proposed models as highlighted in the table, making a new state-of-the-art (SOTA).

\begin{table}[t]
	\centering
	\setlength{\tabcolsep}{3.6pt}\small
	{\begin{tabular}{lcccccc}
		\toprule
		\textbf{Model} &\textbf{VI} &\textbf{ARI} &\textbf{1-1} &\textbf{F1} &\textbf{P} &\textbf{R} \\ 
		\midrule
        \text{BERT$_{base}$}  & 91.7 &74.6 & 80.2 & 33.5 &32.16 & 35.0 \\
        \cdashline{1-7}[0.8pt/2pt]
        \multicolumn{7}{c}{\textit{Ablation study}}\\
        \quad + speaker  &94.0 &81.2 & 84.9 &45.0 &44.7 &45.3\\
        \quad + reference  &94.1 &82.4 &85.6 & 47.4  & 47.4 & 47.4  \\
        \quad + Both &\textbf{94.4} & \textbf{81.8} & \textbf{86.1} & \textbf{52.6} &\textbf{51.0} & \textbf{54.3} \\
		\cdashline{1-7}[0.8pt/2pt]
		\multicolumn{7}{c}{\textit{Aggregation methods}}\\
		\quad w/ max-pooling  &94.1& 80.0 &85.3 & 50.8 & \textbf{52.5} & 49.2  \\
        \quad w/ \texttt{[CLS]} &\textbf{94.4} & \textbf{81.8} & \textbf{86.1} & \textbf{52.6} &51.0 & \textbf{54.3}  \\
        \cdashline{1-7}[0.8pt/2pt]
        \multicolumn{7}{c}{\textit{Layers of Syn-LSTM}}\\
        \quad w/ 1 layer  &\textbf{94.4} & \textbf{81.8} & \textbf{86.1} & \textbf{52.6} &51.0 & \textbf{54.3}   \\
        \quad w/ 2 layers &94.0 & 78.2 &84.6 & 50.4 & 50.9 & 50.0  \\
        \quad w/ 3 layers &94.3 & 79.6 &85.3 & 52.2 & \textbf{51.9} & 52.6  \\
		\bottomrule
	\end{tabular}
	}
	\caption{Results of architecture optimizing experiments.}
	\label{ablation}
\end{table}



\section{Analysis}
\subsection{Architecture Optimizing}
\subsubsection{Ablation Study}
We study the effect of speaker property and reference dependency respectively to verify their specific contribution. We ablate either of the characters and train the model. Results in Table \ref{ablation} show that both speaker property and reference dependency are non-trivial.

\subsubsection{Methods of Aggregation}
At the stage of aggregation heading for context-level representations, we consider the influence of different methods of aggregation, i.e., max-pooling and extraction of \texttt{[CLS]} tokens, the models are trained with the same hyper-parameters. Results in Table \ref{ablation} show \texttt{[CLS]} tokens is a better representation.

\subsubsection{Layers of LSTM}
To determine the optimal depth of the Bi-Syn-LSTM, we do experiments on the number of layers of a Syn-LSTM, also with the same hyper-parameters. According to the results, as shown in Table \ref{ablation}, we put a one-layer Bi-Syn-LSTM for better performance.
\subsection{Prediction Analysis}
To intuitively show and discuss the advantages of the proposed approach, we analyze predictions made by our model and the baseline model (i.e., BERT) in the following aspects.

\begin{figure}[t]
\centering
\begin{subfigure}{.5\linewidth}
  \centering
  \includegraphics[height=5cm]{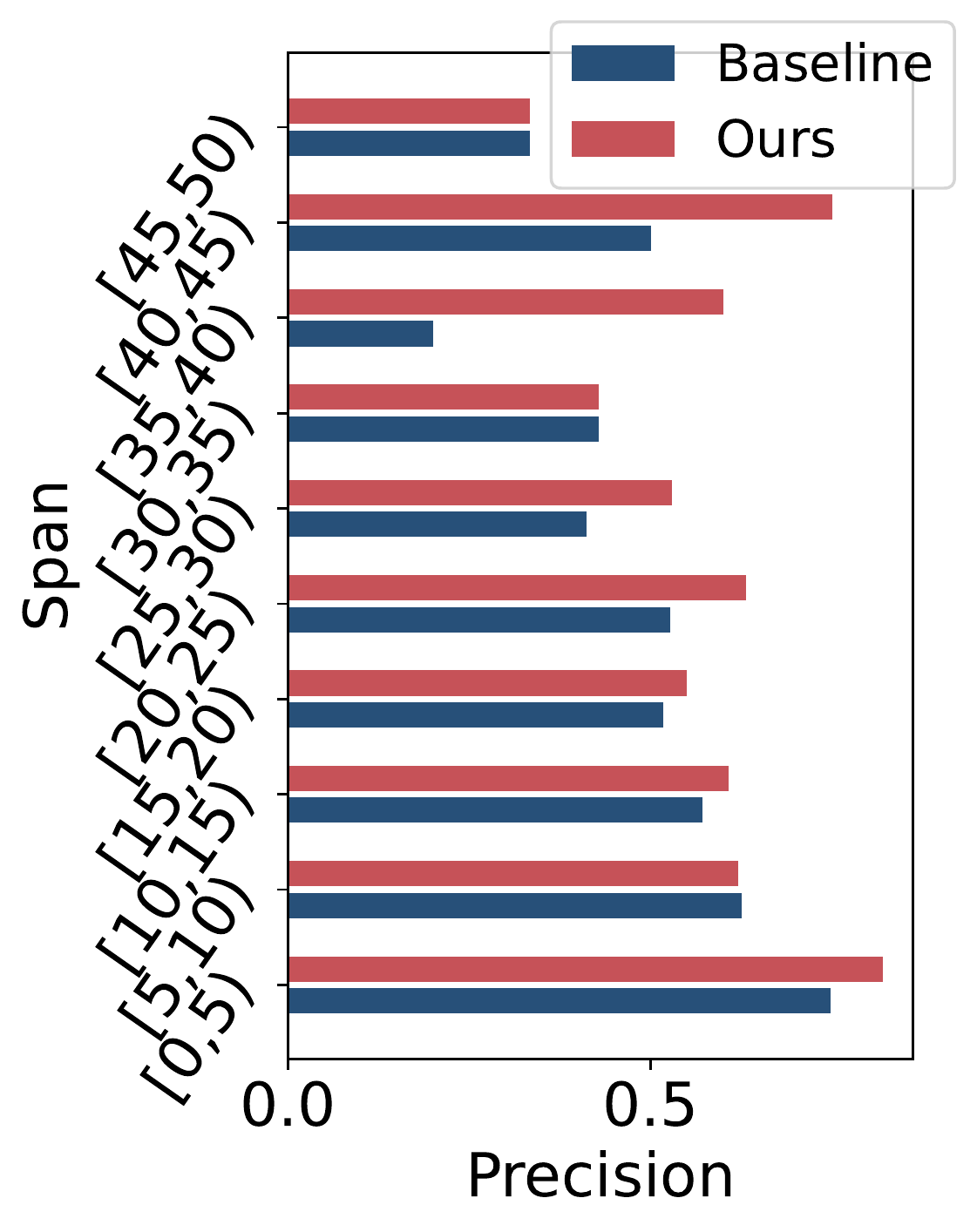}
  \caption{}
  \label{fig:sub1}
\end{subfigure}%
\begin{subfigure}{.5\linewidth}
  \centering
  \includegraphics[height=5cm]{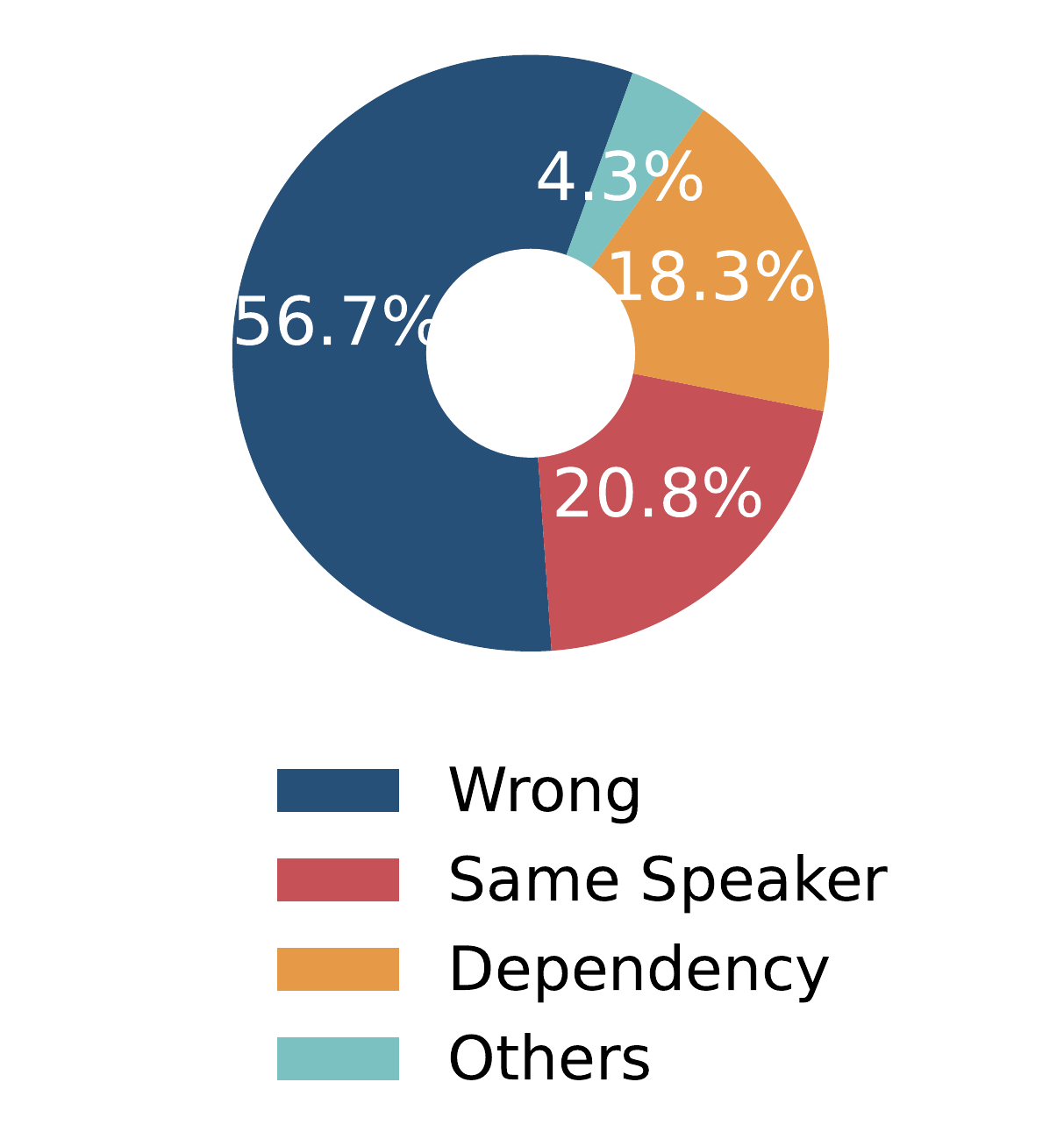}
  \caption{}
  \label{fig:sub2}
\end{subfigure}
\caption{Analysis on (a) Precision on different span lengths. (b) Bad case study.}
\label{fig:test}
\end{figure}

1) We categorize reply-to relationships based on the length of their golden spans (in utterances), and compute the precision of the baseline model and ours. Figure \ref{fig:sub1} shows that our model outperforms baseline by larger margins on links with longer spans (longer than 20 utterances), indicating that our model is more robust on the longer passages.
		

2) We select bad cases of the baseline model to find out how the structure-aware modeling benefits dialogue disentanglement.
We study predictions from our model on these bad cases.
As depicted in Figure \ref{fig:sub2}, the model well solves 43.3\% bad cases.
Our model is observed to correct 20.8\% bad cases whose utterance pairs are from the same speakers, and 18.3\% bad cases whose utterance pairs have a reference.  
As the illustration shows, our model effectively captures the structural features caused by speaker property and reference dependency, thus gaining improvement. 
56.7\% predictions are still wrong. It may suggest deeper inner relationships remain to be studied. 

\subsection{Metrics}
The used metrics are explained and analyzed briefly for a better understanding of model performance in Appendix \ref{sec:appendix1}.

\section{Applications}
Empirically, it is consistent with our intuition that clarifying the structure of a passage helps with reading comprehension. This section studies the potential of dialogue disentanglement by conducting experiments on different tasks and domains.


\subsection{Response Selection}
The dataset of DSTC7 subtask1 \cite{gunasekara-etal-2019-dstc7} is a benchmark of response selection tasks, derived from Ubuntu chatting logs, which is challenging because of its massive scale. As shown in Table \ref{datasets}, it contains hundreds of thousand dialogue passages, and each dialogue has speaker-annotated messages and 100 response candidates.

In the implementation, pre-processed context passages are firstly fed into the trained model for disentanglement to obtain predicted partitions of context utterances.
Then when dealing with the response selection task, we add a self-attention layer to draw attention between utterances within a common cluster in the hope of labels of clusters leading to better contributions to performance.

\subsection{Dialogue MRC}
We also make efforts to apply disentanglement on span extraction tasks of question answering datasets, where we consider multi-party dialogue dataset Molweni \cite{li2020molweni}, a set of speaker-annotated dialogues with some questions whose answers can be extracted from contexts, which is also collected from Ubuntu chatting logs \ref{datasets}.
Because passages in Molweni are brief compared to other datasets we used, utterances tend to belong to the same conversation session through criss-crossed relations. Thus we alternatively leverage labels of reply-to relations from our model, and build graphs among utterances. 

\subsection{Open-domain QA}
As the former two datasets are both extracted Ubuntu IRC chatting logs, we additionally consider an open-domain dataset, FriendsQA \cite{yang-choi-2019-friendsqa}. It contains daily spoken languages from the TV show \textit{Friends} \ref{datasets}. FriendsQA gives QA questions and is handled in the same way as the Molweni dataset.



\begin{table}[hbt]
	\centering
	\setlength{\tabcolsep}{3pt}\small
	{\begin{tabular}{lccc}
		\toprule
		  &\textbf{DSTC-7} &\textbf{Molweni} &\textbf{FriendsQA}\\ 
		\midrule
        \text{Train (dial. / Q)}  & 100,000/-- & 8,771 / 24,682    &973 / 9,791 \\
        \text{Dev (dial. / Q)} &5000/-- &883 / 2,513  & 113 / 1,189\\
        \text{Test (dial. / Q)} &1000/-- &100 / 2,871  & 136 / 1,172\\
        \text{Utterances}  &3-75 & 14 & 173\\
        \textup{Responses} &100 &- &-\\
        \text{Open-domain} &N &N &Y\\
		\bottomrule
	\end{tabular}
	}
	\caption{Statistics of datasets for applications.}
	\label{datasets}
\end{table}

\begin{table}[hbt]
	\centering
	\setlength{\tabcolsep}{3.5pt}\small
	{\begin{tabular}{lcccccc}
		\toprule
		\multirow{2}{*}{\textbf{Model}} &\multicolumn{2}{c}{\textbf{DSTC-7}} & \multicolumn{2}{c}{\textbf{Molweni}} & \multicolumn{2}{c}{\textbf{FriendsQA}} \\
		  &R@1 &MRR &EM &F1 &EM &F1  \\ 
		\midrule
        Public Baseline &- &- & 45.3 &58.0 &45.2 & -- \\
        \text{BERT$_{base}$}  &51.2 &60.9 & 45.7  & 58.8 &45.2 & 59.6 \\
        \quad \text{w/ label}   &51.4 &61.5 & 46.1 &61.7 & 45.2  &60.9\\
        \bottomrule
	\end{tabular}
	}
	\caption{Results of application experiments.}
	\label{qa_res}
\end{table}

Results of the above experiments are presented in Table \ref{qa_res}. It is shown that the disentanglement model brings consistent profits to downstream tasks. Yet, gains on FriendsQA are less impressive, indicating domain limitations to some extent.
Here we only consider naive baselines and straightforward methods for simplicity and fair comparison, which suggests there is still latent room for performance improvement in future work.

\section{Conclusion}
In this paper, we study disentanglement on long multi-party dialogue records and propose a new model by paying close attention to the characteristics of dialogue structure, i.e., the speaker property and reference dependency. 
Our model is evaluated on the largest and latest benchmark dataset Ubuntu IRC, where experimental results show a new SOTA performance and advancement compared to previous work. 
In addition, we analyze the contribution of each structure-related feature by ablation study and the effect of the different model architecture. 
Our work discloses that speaker and dependency-aware structural characters are significant and deserve studies in multi-turn dialogue modeling.

\bibliography{anthology,custom}
\bibliographystyle{acl_natbib}

\appendix

\section{Appendix}

\subsection{Metrics}
\label{sec:appendix1}
The metrics for evaluating the performance of disentanglement are described as follows.

\noindent
\textbf{1) scaled-Variation of Information.} For the two partition $X$ and $Y$ of set $S$, $VI(X;Y) = H(X,Y) - I(X,Y)$, where $H(X,Y)$ is the joint entropy of $X$ and $Y$ and $I(X,Y)$ is the mutual information between $X$ and $Y$, both can be easily calculated from the contingency table. Following previous work\cite{kummerfeld2019large}, VI is scaled to be positive and between 0 and 1. i.e., $1-VI/log_2 (n)$, where n is the number of elements in the set $S$. Thus a bigger number means the two partitions are more similar.

\noindent
\textbf{2) Adjusted Rand Index.} The adjusted Rand index is the corrected-for-chance version of the Rand index \cite{hubert1985comparing}. ARI measures the links between elements under two partitions and indicates how many links lie in the $i$-th part of the predicted partition $X$ and the $j$-th part of the ground truth partition $Y$. Given a contingency table, ARI can be formulated as:
$$\frac{\sum_{ij}C^2_{n_{ij}}-[\sum_i C^2_{a_i}\sum_j C^2_{b_j}]/C^2_{n_{ij}}}{\frac{1}{2}[\sum_i C^2_{a_i}+\sum_j C^2_{b_j}]-[\sum_i C^2_{a_i}\sum_j C^2_{b_j}]/C^2_{n_{ij}}}$$
, where $a_i$ is the summation if row $i$ and $b_j$ is the summation of column $j$. $C$ denotes combinatorial number.

\noindent
\textbf{3) One-to-One Overlap.} 
One-to-one overlap, also called one-to-one accuracy, is calculated as the percentage overlap by pairing up clusters from two partitions to maximize overlap using the methods of max-flow algorithm \cite{elsner-charniak-2008-talking}, indicating how well a whole conversation can be extracted intact.

\noindent
\textbf{4-6) Exact Match.}
Precise, Recall, and F1 score are metrics to measure the exact matching of clusters, where single utterances (clusters only consist of one utterance) are discarded, following previous work.

\begin{figure}[h]
		\centering
		\includegraphics[width=0.5\textwidth]{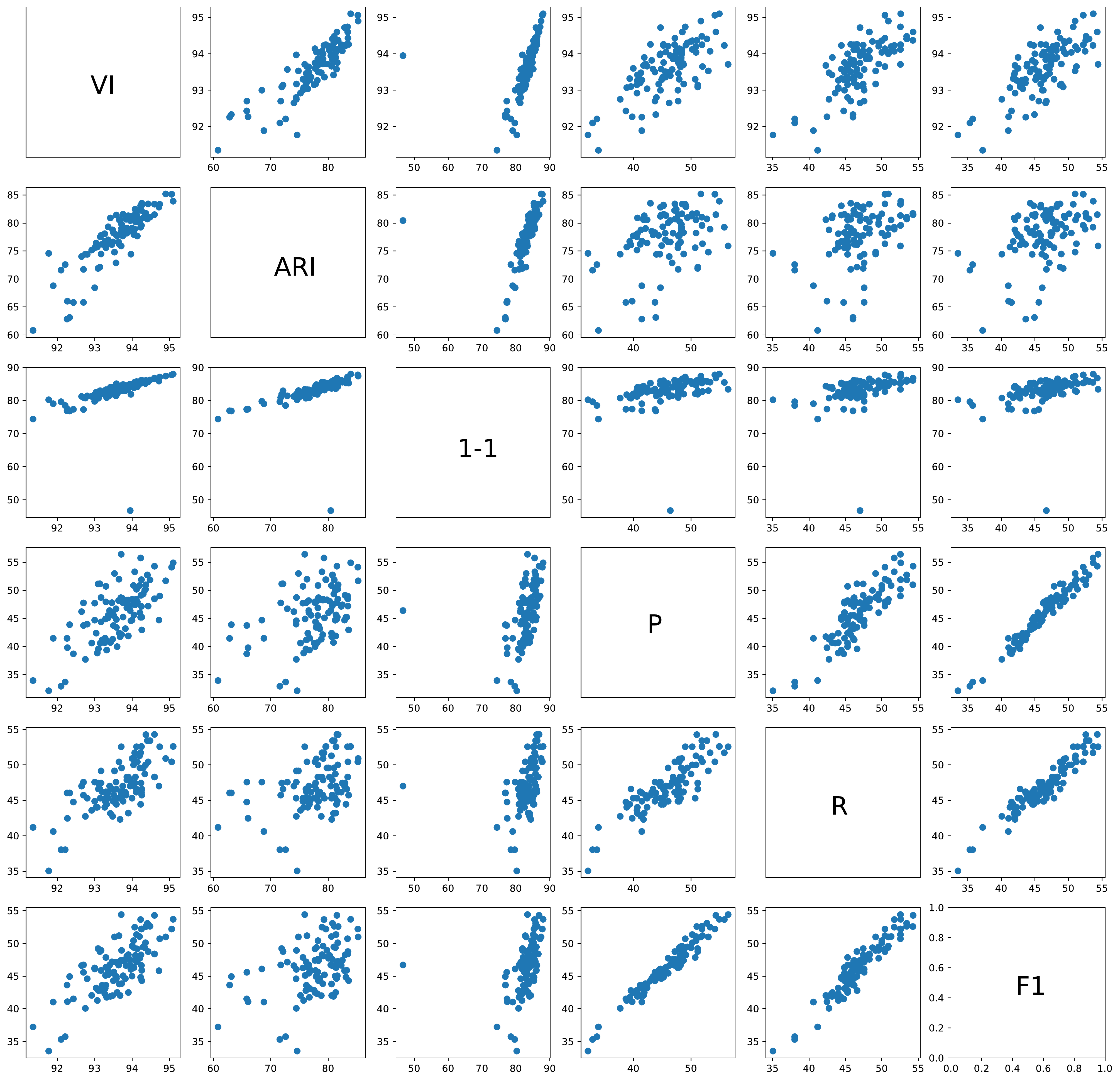}
		\caption{\label{metrics} Scatter plots matrix for metrics. }
\end{figure}

Recently study made efforts to analyze measures \cite{jiang2021dialogue}, where human satisfaction measures are applied on metrics: Normalized Mutual Information (NMI), Adjusted Rand Index (ARI), Shen-F, and F1. Results show that F1 is the most similar to human satisfaction scores, while ARI, NMI, and Shen-F tend to overrate disentanglement results but F1 underrates. Here we present a scatterplot \ref{metrics} based on our experimental results.

\subsection{Syn-LSTM}
As space is limited, we present a complete mathematical representation of Syn-LSTM here \cite{xu2021better}.
\begin{equation}\nonumber
\setlength{\abovedisplayskip}{2pt}
\setlength{\belowdisplayskip}{2pt}
\begin{split}
&\textit{f}_{t} = \textup{sigmoid(W}^{(f)} \textup{x}_{{1}_{t}} + \textup{U}^{(f)} \textup{h}_{t-1} +\textup{Q}^{(f)} \textup{x}_{{2}_{t}} +\textup{b}_{f} \textup{),}\\
&\textit{o}_{t} = \textup{sigmoid(W}^{(o)} \textup{x}_{{1}_{t}} + \textup{U}^{(o)} \textup{h}_{t-1} +\textup{Q}^{(o)} \textup{x}_{{2}_{t}} +\textup{b}_{o} \textup{),}\\
&\textit{i}_{{1}_{t}} = \textup{sigmoid(W}^{(i_1)} \textup{x}_{{1}_{t}} + \textup{U}^{(i_1)} \textup{h}_{t-1} +\textup{b}_{i_1} \textup{),}\\
&\textit{i}_{{2}_{t}} = \textup{sigmoid(W}^{(i_2)} \textup{x}_{{2}_{t}} + \textup{U}^{(i_2)} \textup{h}_{t-1} +\textup{b}_{i_2} \textup{),}\\
&\textit{c}_{{1}_{t}} = \textup{tanh(W}^{(k)} \textup{x}_{{1}_{t}} + \textup{U}^{(k)} \textup{h}_{t-1} +\textup{b}_{k} \textup{),}\\
&\textit{c}_{{2}_{t}} = \textup{tanh(W}^{(p)} \textup{x}_{{2}_{t}} + \textup{U}^{(p)} \textup{h}_{t-1} +\textup{b}_{p} \textup{),}\\
&\textit{c}_{t} = \textit{f}_{t} \textup{$\odot$}\textit{c}_{t-1} + \textit{i}_{{1}_{t}} \textup{$\odot$}\textit{c}_{{1}_{t}} + \textit{i}_{{2}_{t}} \textup{$\odot$}\textit{c}_{{2}_{t},}\\
&\textit{h}_{t} = \textit{o}_{t} \textup{$\odot$} \textup{tanh(c}_{t}\textup{),}\\
\end{split}
\label{synfull}
\end{equation}
,where $x_{1_t}$ and $x_{2_t}$ are inputs.$c_{t-1}, c_t$ denote former and current cell states. $h_{t-1}$ is former hidden state. $W, U, b$ are learnable parameters. $f_t, o_t, i_{1_t}, i_{2_t}$ are forget gate, output gate and two input gates.

\end{document}